\documentclass{article}
\usepackage{ijcai22}

\usepackage{times}
\usepackage{soul}
\usepackage{url}
\usepackage[hidelinks]{hyperref}
\usepackage[utf8]{inputenc}
\usepackage[small]{caption}
\usepackage{graphicx}
\usepackage{amsmath}
\usepackage{amsthm}
\usepackage{booktabs}
\usepackage{algorithm}
\usepackage{helvet}  
\usepackage{courier}  
\usepackage[switch]{lineno}
\usepackage{multirow}
\usepackage{todonotes}
\usepackage{amsmath}
\usepackage{enumerate}
\usepackage{balance}
\usepackage{booktabs}
\usepackage{amssymb}
\usepackage{amsfonts}
\usepackage{mathtools}
\usepackage{comment}
\usepackage{subfigure}
\usepackage{latexsym}
\usepackage{color}
\usepackage{algorithmicx}  
\usepackage{algpseudocode}  
\usepackage{setspace}
\usepackage{paralist}

\usepackage{microtype}
\usepackage{booktabs}
\usepackage{amsmath,amsthm,amsfonts,amssymb,bm}
\usepackage{mathrsfs}
\usepackage{amsmath}
\usepackage{algorithm}  
\usepackage{algpseudocode}  
\usepackage{booktabs}
\usepackage{enumitem}
\usepackage{amsthm}
\usepackage{amssymb}
\newcommand{\tabincell}[2]{\begin{tabular}{@{}#1@{}}#2\end{tabular}}
\theoremstyle{definition}
\usepackage{multirow}
\usepackage{verbatim}
\usepackage{etoolbox}
\usepackage{hyperref}
\usepackage{subfigure}
\usepackage{multirow}
\usepackage{siunitx}
\usepackage{cleveref}
\crefname{section}{§}{§§}
\Crefname{section}{§}{§§}

\setitemize[1]{leftmargin=10pt,itemsep=0.5pt,partopsep=0pt,parsep=0.5pt,topsep=0pt}

\title{``My nose is running.” ``Are you also coughing?”:\\ Building a Medical Diagnosis Agent with Interpretable Inquiry Logics}

\author{
Wenge Liu$^{1}$\thanks{Equal contribution.}, Yi Cheng$^{2}$\footnotemark[1], Hao Wang$^{3}$, Jianheng Tang$^{4}$, \\Yafei Liu$^{5}$, Ruihui Zhao$^{5}$, Wenjie Li$^{2}$, Yefeng Zheng$^{5}$, Xiaodan Liang$^{1}$\thanks{Corresponding author.}\\
\affiliations
$^1$ Sun Yat-sen University\quad
$^2$ Hong Kong Polytechnic University \quad
$^3$ Rutgers University\quad\\
$^4$ Hong Kong University of Science and Technology\quad
$^5$Tencent Jarvis Lab\\
    \tt \{kzllwg,hoguewang,sqrt3tjh,xdliang328\}@gmail.com\\
    \tt alyssa.cheng@connect.polyu.hk, cswjli@comp.polyu.edu.hk \\
 \tt \{davenliu,zacharyzhao,yefengzheng\}@tencent.com\\
}

\let\oldequation\equation
\let\oldendequation\endequation
\renewenvironment{equation}{\linenomathNonumbers\oldequation}{\oldendequation\endlinenomath}
\begin{document}
\maketitle
\begin{abstract}

With the rise of telemedicine, the task of developing Dialogue Systems for Medical Diagnosis (DSMD) has received much attention in recent years. Different from early researches that needed to rely on extra human resources and expertise to help construct the system, recent researches focused on how to build DSMD in a purely data-driven manner. However, the previous data-driven DSMD methods largely overlooked the system interpretability, which is critical for a medical application, and they also suffered from the data sparsity issue at the same time. In this paper, we explore how to bring interpretability to data-driven DSMD. Specifically, we propose a more interpretable decision process to implement the dialogue manager of DSMD by reasonably mimicking real doctors' inquiry logics, and we devise a model with highly transparent components to conduct the inference. Moreover, we collect a new DSMD dataset, which has a much larger scale, more diverse patterns and is of higher quality than the existing ones. The experiments show that our method obtains 7.7\%, 10.0\%, 3.0\% absolute improvement in diagnosis accuracy respectively on three datasets, demonstrating the effectiveness of its rational decision process and model design. Our codes and the GMD-12 dataset are available at \url{https://github.com/lwgkzl/BR-Agent}.

\end{abstract}

\section{Introduction}

Due to the widespread shortage of medical resources, millions of patients around the world are facing the delay of disease diagnosis and therapy. 
To automate the process of medical consultations and relieve the therapeutic stress, it has been researched for decades on how to develop Dialogue Systems for Medical Diagnosis (DSMD). Typically, a DSMD needs to keep inquiring about the patient's symptoms in multiple turns until a preliminary diagnosis can be confidently rendered in the final turn. 
In this line of research, early works needed to rely on extra human resources and expertise to help construct the system, such as for feature engineering and rule designing \cite{shortliffe1974mycin,PopleMM75,milward2003ontology}. 
It was not until 2018 that Wei et al. \shortcite{wei2018task} proposed the first DSMD dataset collected from an online healthcare community, and started the research trend on developing DSMDs in a purely data-driven manner that does not require any extra labour \cite{xu2019end,xia2020generative,liao2020task,he2020fit,zhao2020graph,liu2021heterogeneous,liu2020meddg,lin2021graph}. 
Before this dataset, previous researches needed to simulate the patient's situations to test the system performance, rather than used data directly collected from the cases of real patients. 

\begin{figure}[t]
\centering
  \includegraphics[width=.9\linewidth]{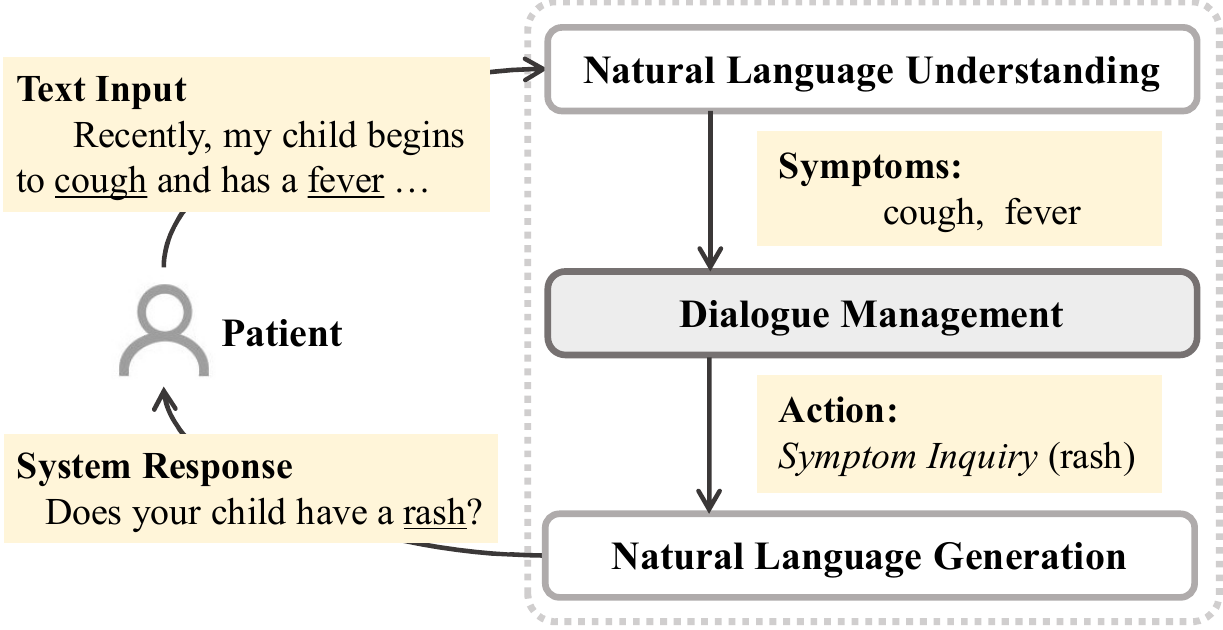}
\caption{DSMD components and their functions.}
\label{fig:framework}\vspace{-2mm}
\end{figure}

However, despite avoiding the dependence on extra human resources to construct the system, these data-driven methods largely overlooked the system interpretability, which is critical for a medical application. As shown in Fig. \ref{fig:framework}, a DSMD typically consists of three components and the Dialogue Management (DM) module is the most central part. 
It is responsible for selecting the next action for the system, either querying a symptom or giving the final diagnosis. 
In other words, suppose there are $n$ types of symptoms and $m$ types of diseases; DM needs to conduct classification among the $n$+$m$ possible actions. 
In the previous researches on data-driven DSMD, they implemented the DM with black-boxed neural networks and directly generated the probability distribution of all the $n$+$m$ possible actions, i.e., conducting disease inference and symptom selection simultaneously. 
Such a decision process is very weakly interpretable and their model components also lack enough transparency.

To the best of our knowledge, \textbf{this is the first work that explores how to bring interpretability to data-driven DSMD.} 
It needs to be clarified that the interpretability we aim to achieve here is to design an \emph{intrinsically} interpretable method for DSMD, rather than use some post-hoc methods to explain a trained black-boxed model \cite{molnar2018interpretable}. 
To this end, we propose a more interpretable two-stage decision process to implement the DM of DSMD, by reasonably mimicking real doctors' consultation logics. 
We argue that it is more interpretable to conduct disease inference and symptom selection \emph{successively}, rather than \emph{simultaneously} as in the previous researches. 
Specifically, at each dialogue turn, the system should first infer the patient's possible diseases based on the current symptom information.  
If the most suspected disease reaches a confidence threshold, it would give the diagnosis result; otherwise, it would further inquire a symptom according to the disease estimation. 
It is just as in real consultation scenarios, where a doctor would only inquire about a symptom due to suspicion of particular diseases, rather than directly based on the already-known symptoms. 
Moreover, for symptom selection, we also summarize that there are two kinds of selection logics: a symptom is queried either to \emph{ensure} the suspicion of one disease or to \emph{distinguish} similar diseases. 

Correspondingly, we propose a model with highly transparent components to conduct the above decision process, named Bayesian Reinforced Agent (BR-Agent). It includes a BayesNet for disease inference and two matrices that respectively mimic two kinds of symptom selection logics, which are controlled by a logic switcher. 
Their parameters are all practically meaningful. For instance, the parameter in the BayesNet is either the prior probability of a disease or the conditional probability of a symptom given diseases. 

For the lack of turn-level supervision labels, BR-Agent is optimized via long-term reinforced rewards that consider symptom recall and diagnosis accuracy. 
The BayesNet in BR-Agent is also trained end to end with the other components via the gradient from RL, which is different from the learning paradigm of other BayesNet applications in the medical field. 
In the previous researches, the BayesNet parameters were usually determined with the help of expert knowledge or using statistical estimation methods \cite{lincoln1967medical,kahn1997construction,wang1999computer,mcgeachie2009integrative,flores2011incorporating}. 
Even when combining BayesNets with other neural networks in a deep learning fashion, the parameters of BayesNets were also learned separately using more traditional methods, rather than trained end to end with the other components as we do \cite{chen2020towards,kim2021perspective}.  
For instance, Chen et al. \shortcite{chen2020towards} implemented a BayesNet in combination with hierarchical CNNs. They used the gradient mechanism to train the CNN part, while the BayesNet parameters were just estimated by counting the feature occurrences in the dataset.

Our proposed method is not only interpretable, but it also demonstrates very competitive performance due to its rational decision process and model design. It \textbf{exceeds the previous state-of-the-art by a large margin in diagnosis accuracy}, with an absolute improvement of  7.7\%, 3.0\%, 10.0\% respectively on three datasets.
Since the existing DSMD data is still very limited, \textbf{we also collect a new dataset, GMD-12}, which has a larger scale, more diverse patterns, and is of higher quality than the previous datasets. 
Its scale is more than three times of the existing datasets in terms of dialogue number. The number of disease and symptom types in GMD-12 are also much more diverse, so new patterns can be observed from it. 
Moreover, our data is obtained from collaborating hospitals and revised by clinical experts, while the previous datasets were directly crawled from telemedicine websites, so GMD-12 is more professional and of higher quality.

\section{Preliminaries}
As shown in Fig. \ref{fig:framework}, a DSMD typically consists of a Natural Language Understanding (NLU) module, a DM module, and a Natural Language Generation (NLG) module. 
At each turn, based on the currently-known symptoms extracted through NLU, DM either chooses a disease as the diagnosis result and finishes the dialogue, or selects one more symptom to query. The output of DM is then transformed into natural language through NLG. 
Our work, as well as the previous DSMD researches, \textbf{only focus on developing the DM part}, because, for privacy issues, the released versions of DSMD datasets do not contain the original dialogue content between patients and doctors. They only contain processed structured data to test DM alone, with no need for NLU or NLG. 
Besides, the construction of NLU and NLG is relatively simple. For instance, NLU can be addressed by designing a set of regular expressions, as patients' expression patterns of the symptoms are relatively fixed; NLG can be implemented with template-based methods, since language diversity and vividness are less important in DSMD. 

The task of DM in DSMD can be formally defined as below. 
Suppose there are $N$ types of symptoms and $M$ types of diseases. 
The input of DM at the $t$-th turn is $\bm s_{t}  \in \mathbb{R}^{N}$, which represents the system's current knowledge of the patient's symptoms.  
Each dimension of $\bm s_{t}$ corresponds to one symptom and takes value from $\{1,-1,0\}$, respectively standing for \emph{positive}, \emph{negative}, and \emph{uncertain}. 
Given $\bm s_t$, DM either selects one more symptom to query, or chooses a disease as the diagnosis result. Thus, there are overall $N+M$ possible actions to choose from.

\section{Methodology}
\begin{figure}[t]
\large
\centering
  \includegraphics[width=\linewidth]{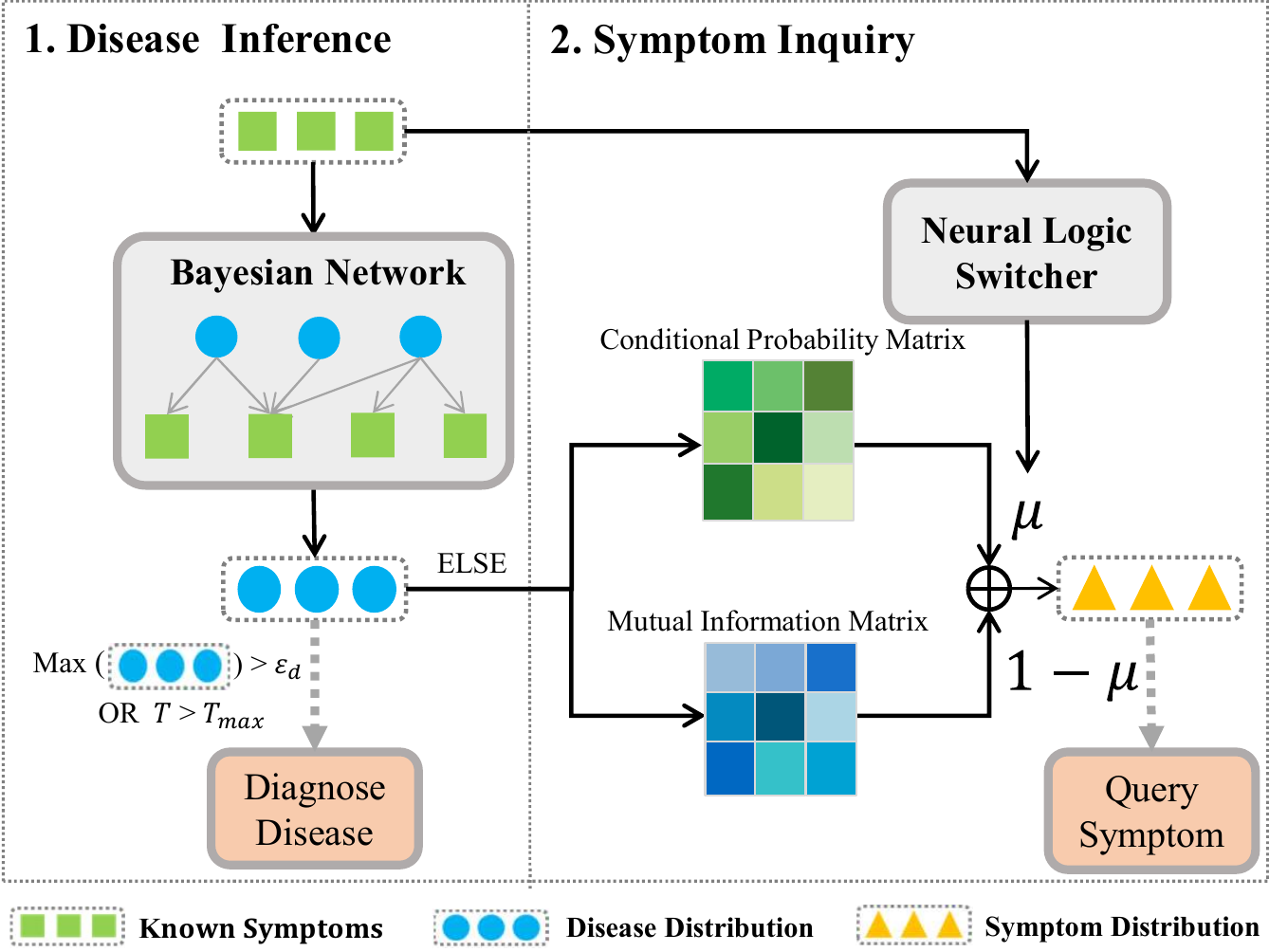}
\caption{Overview of our proposed dialogue manager for DSMD. }
\label{fig:model}
\end{figure}

Fig. \ref{fig:model} presents an overview of our proposed dialogue manager for DSMD, named BR-Agent. 
At the $t$-th turn, given symptom information $\bm s_{t}$, a BayesNet is first adopted to infer the patient's disease $\bm P_D$.
If the probability of the most suspected disease (i.e., the maximum value in $\bm P_{D}$) is larger than the threshold $\varepsilon_{d}$ or $t$ reaches the maximum turn number $T_{max}$, DM would generate the diagnosis result.
Otherwise, we would further predict which symptom most needed to query in the next response $\bm P_S$, using a neural logic switcher and two matrices that simulate two different inquiry logics. 
Below we will more detailedly describe the two parts of BR-Agent, \emph{disease inference} and \emph{symptom inquiry}. Then, we will illustrate how we train BR-Agent with reinforcement learning.

\subsection{Disease Inference}\label{sec:di}
The structure of the BayesNet in BR-Agent models the relationship between diseases and symptoms by a directional bipartite graph $\mathcal G$=$(\mathcal V,\mathcal E)$. The node set $\mathcal V$ consists of disease nodes $\mathcal{D}$=$\{D_i |i=1,2,...,M\}$ and symptom nodes $\mathcal{S}$=$\{S_j|j=1,2,...,N\}$. The edge set $\mathcal E$ is constructed by counting co-occurrence times of each disease-symptom pair in the corpus. Concretely, the disease node $D_i$ points to the symptom node $S_j$ if the co-occurrence number of $D_i$ and $S_j$ is greater than a given threshold $\epsilon_e$. The parent set of $S_j$ (the set of all nodes pointing to $S_j$) is denoted as $\mathtt{Parents}(S_j)$, which represents all possible diseases causing this symptom.

For parameter learning, we first initialize the BayesNet parameters by referring to the Bayesian parameter estimation method in ~\cite{chen2020towards}, and then fine-tune them during the end-to-end RL training with other components. 
Specifically, there are two types of parameters in our BayesNet $\bm \theta_{BN}$. 
One is $\bm P(D_i)$, the prior probability of disease $D_i$ $(i=1,$ $2,..,M)$. It is initialized with the prevalence of the disease in the dataset. 
The other is the probability of a symptom conditioned on related diseases $\bm P(S_i|D_i^+, D_i^-)$, where $D_i^+$ is the set of diseases that the patient has and $D_i^-$ is the one that he/she does not have; they satisfy that $\mathtt{Parents}(S_i)$ is their disjoint union, i.e.,  $\mathtt{Parents}(S_i)=D_i^+ \sqcup D_i^-$. 
In our experimental datasets, the diagnosis of a patient only consists of one disease, and we assume that the patient does not have any other unmentioned diseases. Thus, if $|D_i^+|=1$ only has one element, we can estimate this probability based on the co-occurrence relationship of the symptom and the disease. For the other cases, we initialize them with 0.5 as a random guess. 
Specifically, the initial value of parameter $\bm P(S_i|D_i^+,D_i^-)$ is calculated as: 
\begin{equation}
    \bm{P}(S_i|D_i^+, D_i^-) =\left\{
            \begin{array}{ll}
                {n(S_i,D_i^+)}/{n(D_i^+)} & \text{if } {|D_i^+|=1}\\
                0.5 & \text{else.}\\
            \end{array} \right. \notag
\end{equation}
where $n(D_i^+)$ is the number of patients in the dataset who has the disease in $D_i^+$; $n(S_i,D_i^+)$ is the number of dialogues where the patient has symptom $S_i$, the disease in $D_i^+$ and does not have diseases $D_i^-$.

After initialization, all the BayesNet parameters $\bm\theta_{BN}$ will be further fine-tuned during the end-to-end training with other DM components. 
Given the BayesNet and a set of observed positive symptoms $S^+$ and negative symptoms $S^-$ derived from $\bm s_{t}$, we 
adopt the Variable Elimination (VE) algorithm~\cite{VE} to calculate the disease distribution 
$\bm P_{D}= \bm P(\mathcal{D} | S^+, S^-)=\mathtt{VE}(S^+, S^-;\bm \theta_{BN}).$
Essentially, given the observed symptoms, VE marginalizes out the unobserved symptoms $S^u= \mathcal{S} \setminus (S^+\cup S^-)$, and computes the disease distribution $\bm P_D$ conditioned on the observed symptoms. 
Since the calculation process of VE is differentiable, the BayesNet parameters $\bm \theta_{BN}$ can be updated using the gradient on $\bm P_{D}$ during the end-to-end training with other model components in dialogue management. More details on the RL training process is provided in \Cref{sec:e2e}.

\subsection{Symptom Selection}\label{sec:ss}

For symptom selection, we utilize two matrices to simulate two types of human doctors' inquiry logic: a conditional probability matrix $\bm M_c \in \mathbb{R^{M\times N}}$, to ensure suspicion of a specific disease, and the mutual information matrix $\bm M_m \in \mathbb{R^{M\times N}}$, to distinguish two similar diseases. 
The element $\bm M_c(i,j)$ in the conditional probability matrix indicates the probability of symptom $S_{j}$ given the presence of disease $D_{i}$:
\begin{equation}
    \bm M_c(i,j) = \bm P(S_{j}|D_{i}) = \frac{n(D_i,S_j)}{\sum_{k=1}^{N} n(D_i,S_k)},\notag
\end{equation}
where $n(i,j)$ is the number of dialogues where the patient has symptom $S_{j}$ and disease $D_{i}$. 

The element $\bm M_m(i,j) $ in the mutual information matrix measures the mutual information between disease $D_{i}$ and symptom $S_{j}$. Specifically, it is calculated as
\begin{align*}
\bm M_m(i,j) = I(D_i;S_j)&= \sum_{k_{1}}\sum_{k_{2}}\bm P(D_i=k_{1},S_j=k_{2})\\
&\cdot \log\left( \frac{\bm P(D_i=k_{1},S_j=k_{2})}{\bm P(D_i=k_{1}) \bm P(S_j=k_{2})} \right) ,  
\end{align*}
where $k_{1}$ and $k_{2}$ are the sets of possible values for disease $D_{i}$ and symptom $S_{j}$, respectively. In our case, $k_1, k_2 \in \{0,1\}$, where 0 represents negative and 1 positive. 

The condition probability matrix tends to query high-frequent symptoms that co-occur with suspected diseases. 
However, some symptoms can be common for many diseases, such as the symptom ``fever'', and thus cannot be used to distinguish them. 
On the other hand, the mutual information measures the connection between $D_i$ and $S_j$, which indicates whether $S_j$ can be used to distinguish $D_i$ from other diseases, so it can query a low-frequent symptom as long as it has a stronger connection with the suspected disease than others. Both matrices are normalized along the row axis.

To determine which symptom inquiry logic is more suitable for the current turn, we adopt the Multi-Layer Perceptron (MLP) to determine a weight factor $\mu \in [0,1]$ to infuse the prediction results from two matrices. Specifically, $\mu$ and $\bm P_S$ are calculated as:
\begin{align*}
    \mu &= \text{Sigmoid}(\text{MLP}(\bm s_t; \bm P_{D})), \\
    \bm P_S  &= \mu \bm P_D \cdot \bm M_c + (1-\mu) \bm P_D \cdot \bm M_m. 
\end{align*}

\subsection{RL Training Framework}\label{sec:e2e}
Our model is trained end to end with reinforcement learning, using the Advantage Actor Critic (A2C) algorithm \cite{mnih2016asynchronous}. 
The A2C algorithm includes an \emph{actor} to generate actions, an \emph{environment} to update the current state based on the action and generate rewards, and a \emph{critic} that learns a value function to evaluate the quality of the current state through the given reward. This value function is leveraged to train the actor by estimating  the ``advantage'', that is, the difference between the estimated reward and the actual reward received after action $a_t$. 
In our work, the state corresponds to $\bm s_t$, the current knowledge of symptom information, and the actor is just our dialogue manager.
The environment is a user simulator. It has access to all the patient's symptom information from the dataset, so it could answer the DM's symptom inquiry by updating the state. To train the critic, it gives rewards to correct diagnosis results and penalties to mistaken diagnosis and the symptom inquiries that  get negative answers. For the critic, we implement a multi-layer perceptron to calculate the value function $v(\bm s_{t};\bm \theta_{v})$, and the advantage is measured with the temporal difference error $\delta_t =r_t+\gamma v(\bm s_{t+1};\bm \theta_v)-v(\bm s_t;\bm \theta_v)$, where $\gamma$ is a hyper-parameter, denoting the discount rate. The parameters of the actor $\bm \theta_{\pi}$ and the critic $\bm \theta_{v}$ are updated by:

\begin{align*}
    \bm \theta_{\pi} \leftarrow \bm \theta_{\pi} & + \beta_1 \cdot \delta_t \cdot \nabla_{\bm \theta_{\pi}}\ln \pi(a_{t}|\bm s_{t};\bm \theta_{\pi} ) \\ 
     & + \beta_2 \cdot H(\pi(a_{t}|\bm s_{t};\bm \theta_{\pi}) \\ 
    \bm \theta_{v}  \leftarrow \bm \theta_{v} & + \alpha \cdot \delta_t \cdot \nabla_{\bm \theta_v}v(\bm s_t;\bm \theta_v),
\end{align*}
where $\beta_1$, $\beta_2$ and $\alpha$ are hyper-parameters and $H(\cdot)$ is an entropy regularization term to encourage policy exploration.
In this way, the parameters of the BayesNet in the actor are also fine-tuned end to end with other components.

\section{Dataset Construction}

Before this work, there are two public DSMD datasets, Muzhi~\cite{wei2018task} and Dxy~\cite{xu2019end}. 
As shown in Table \ref{tab:comparison_statistics}, the scales of the two public datasets are relatively small. Muzhi contains 710 dialogues, 4 pediatric diseases and 66 symptoms, while Dxy contains 527 medical diagnosis dialogues, 5 diseases and 41 symptoms. Both of them directly crawled the data from online healthcare communities. 

To alleviate the data sparsity issue for DSMD, we further collect a new dataset, named GMD-12. We use the medical records 
from several collaborating hospitals as the data source. The hospitals had already asked the related patients' permission to use their medical records for academic research, with signature confirmation. These offline records are of higher quality than the data directly crawled from the online communities, as they are guaranteed to be real patients' cases given diagnosis by  professional doctors. 

We were first given access to 17,000 medical records, from which we selected the 12 most frequent diseases to include in the dataset. The 118 symptoms were determined accordingly with the help of collaborating clinicians.
Then, we extract the symptom and disease information from the collected records, using an enterprise-level medical entity extraction system trained on a large-scale electronic health record corpus. 
After that, all the extracted symptom and disease were manually normalized to its corresponding terminology on SNOMED CT\footnote{https://www.snomed.org/snomed-ct} and checked by three domain experts.

As shown in Table \ref{tab:comparison_statistics}, besides the larger sample size, our GMD-12 also has more diverse diseases and symptom types, so new patterns can be observed from it. 
To evaluate the data quality of the three datasets, three doctors with more than two-year consultation experience are asked to assess the \emph{rationality score} of 200 samples from each dataset. The rationality score is rated on a 0-2 scale (2 for the best), representing whether the listed symptoms in the given sample can support its diagnosis result. We can see from Table \ref{tab:comparison_statistics} that GMD-12 achieves the highest rationality score, demonstrating that it has better quality from the professional perspective. 

\begin{table}[t]
\centering
\footnotesize
\begin{tabular}{l|ccc}
\specialrule{1.2pt}{1pt}{1pt}
\textbf{Dataset}   & \textbf{Dxy} & \tabincell{c}{\textbf{Muzhi}}& \tabincell{c}{\textbf{GMD-12} (Ours)} \\
\specialrule{1.2pt}{0pt}{0.5pt}
\# Dialogues               & 527   & 710   & \textbf{2,374} \\
\# Disease Types               & 5     & 4   & \textbf{12}  \\ 
\# Symptom Types               & 41    & 66   & \textbf{118} \\
\# Avg. Symptoms/Patient      & 4.74  & \textbf{5.59}   & 5.55 \\
\specialrule{1pt}{0.8pt}{0.8pt}
Rationality Score       & 1.54  & 1.60   & \textbf{1.65}  \\
\specialrule{1.2pt}{0.8pt}{1pt}
\end{tabular}
\caption{Statistics of three DSMD datasets. ``Avg. Symptoms/Patient'' represents the average number of symptoms that one patient has in the dataset. ``Rationality Score'' represents the professional evaluation of the data quality (on a 0-2 scale; 2 for the best).}
\label{tab:comparison_statistics}
\end{table}

\section{Experiments}
\subsection{Experimental Setup}
Our experiments are conducted on two public datasets, Muzhi and Dxy, and our newly-constructed GMD-12 dataset. 
Each sample contains simplified structured data, including the patient's disease and his/her symptoms, with no need of NLU and NLG. 
The symptoms are labelled as either ``explicit'' or ``implicit''. The explicit ones are fed to the DM in the first dialogue turn, and the implicit ones can be quried in the following turns. 
A user simulator that has access to all the symptom information is implemented to interact with the DM and answer its symptom inquiry. 

\subsection{Automatic Evaluation}
\label{sec:setup}
Our automatic metrics include \emph{diagnosis accuracy} and \emph{symptom recall}.
Diagnosis accuracy is the main focus of the DSMD task, i.e., correctly diagnosing the disease. 
Symptom recall is the average proportion of the patient's symptoms that are successfully queried by the dialogue agent. It measures the agent's efficiency in collecting patients' information. 


\paragraph{Comparison with Baselines.}
The compared baselines are as follows. 
\textbf{Basic DQN}~\cite{wei2018task} applies RL to DSMD with a deep Q-network.
\textbf{A2C-GCN}~\cite{kipf2016semi} adopts the A2C algorithm similar to our method, but its actor is implemented with Graph Convolutional Networks (GCN). 
\textbf{Sequicity}~\cite{lei_etal_2018_sequicity} uses the sequence-to-sequence architecture for task-oriented dialogue systems, optimized with RL. 
\textbf{KR-DS}~\cite{xu2019end} and \textbf{GAMP}~\cite{xia2020generative} are two state-of-the-art models for DSMD. 
The evaluation results 
are presented in Table~\ref{table:cmpAbl}. 

We can see that BR-Agent obtains 8\%, 3\%, 10\% absolute improvement in diagnosis accuracy and 14\%, 54\%, 14\% improvement in symptom recall respectively on the Dxy, Muzhi, and GMD-12 datasets. 
It indicates that BR-Agent reasonably models the human doctors' inquiry logics and can more precisely conduct disease diagnosis by collecting more effective symptoms. 
The improvement of diagnosis accuracy on GMD-12 is the largest, verifying better scalability of BR-Agent for larger datasets. 
On the Muzhi dataset, though BR-Agent can effectively recall more symptoms, the improvement of diagnosis accuracy is still relatively small. 
It is probably because the diseases in Muzhi are very similar and difficult to distinguish (e.g., \emph{children’s bronchitis} and \emph{infantile diarrhoea}).

\begin{table}[t]
\footnotesize
\centering
\begin{tabular}{l|cc | cc | cc}
\specialrule{1.2pt}{1pt}{1pt}
\multirow{2}{*}{\bf \tabincell{c}{\textbf{Methods}}} & \multicolumn{2}{c|}{\textbf {Dxy}} & \multicolumn{2}{c|}{\textbf {Muzhi}} & \multicolumn{2}{c}{\textbf {GMD-12}}\\
&Acc.  & Rec. &Acc.  & Rec. &Acc.  & Rec.\\
\specialrule{1.2pt}{0pt}{0.5pt}
Basic DQN &0.731&0.245 &0.65 &0.04&0.62&0.05 \\
A2C-GCN &0.740&0.169 &0.69&0.09&0.72&0.36 \\
Sequicity &0.285&0.246&-&-&-&- \\
KR-DS &0.740&0.342 &0.73&0.13&0.69 &0.21 \\
GAMP &0.769&0.170 &0.73&-&-&- \\
\specialrule{0.8pt}{0.5pt}{0.5pt}
BR-Agent &\textbf{0.846}&\textbf{0.486} &\textbf{0.76}&\textbf{0.670}&\textbf{0.82}&\textbf{0.50} \\
\specialrule{1.2pt}{1pt}{1pt}
\end{tabular}
\caption{The diagnosis accuracy and symptom recall of BR-Agent and other baselines on the Dxy, Muzhi, and GMD-12 datasets.}
\label{table:cmpAbl}
\end{table}
\begin{table}[t]
\footnotesize
\centering
\begin{tabular}{l|cc | cc | cc}
\specialrule{1.2pt}{1pt}{1pt}
\multirow{2}{*}{\bf \tabincell{c}{\textbf{Methods}}} & \multicolumn{2}{c|}{\textbf {Dxy}} & \multicolumn{2}{c|}{\textbf {Muzhi}} & \multicolumn{2}{c}{\textbf {GMD-12}}\\
&Acc.  & Rec. &Acc.  & Rec. &Acc.  & Rec.\\
\specialrule{1.2pt}{0pt}{0.5pt}
BR-Agent &\textbf{0.846}&\textbf{0.486} &\textbf{0.76}&0.67&\textbf{0.82}&\textbf{0.50} \\
w/o mutual. &0.837&0.382  &0.75&\textbf{0.70}&0.81&0.44\\
w/o cond.   &0.817& 0.437  &0.74&0.57&0.80 &0.43\\
w/o matrices   &0.760&0.284  &0.73&0.29&0.74 &0.17\\
w/o BayesNet  &0.721&0.161  &0.67 &0.06&0.65&0.22 \\
\specialrule{1.2pt}{1pt}{1pt}
\end{tabular}
\caption{Ablation study of BR-Agent, including the results of removing the mutual information matrix, the conditional probability matrix, both matrices, and the BayesNet.}
\label{table:Abl}
\end{table}

\begin{table}[t]
\centering
\footnotesize
\begin{tabular}{l |cc|cc}
\specialrule{1.2pt}{1pt}{1pt}
\textbf{Methods} & \textbf{Inquiry} & $\bm \kappa$ & \textbf{Diagnosis} & $\bm \kappa$\\
\specialrule{1.0pt}{0.5pt}{0pt}
Transparent-A2C  & 2.35 & 0.82 & 2.78 & 0.69\\
Retrieval & 2.52 & 0.70 & 1.22 & 0.80\\
BR-Agent & \textbf{4.30} & 0.78 & \textbf{4.39} & 0.75\\
\specialrule{1.2pt}{1pt}{1pt}
\end{tabular}
\caption{Human evaluation results of whether the model's symptom inquiry and the diagnosis actions are rational from professional perspectives, along with the Cohen's kappa between two annotators  (on a 1-5 scale; 5 for the best).}
\label{table:human}
\end{table}

\begin{figure}[t]
\centering
  \includegraphics[width=\linewidth]{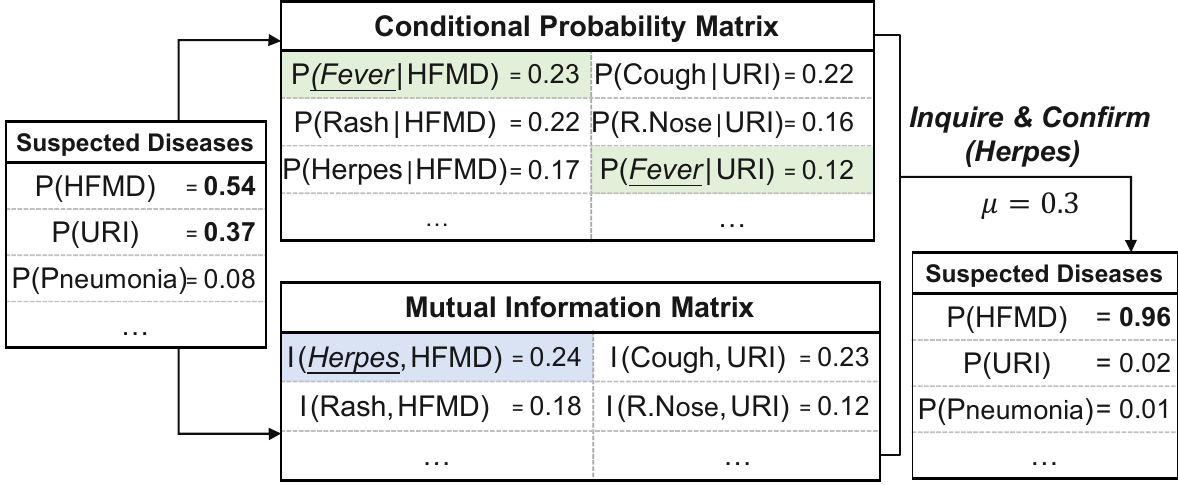}
\caption{The conditional probability matrix tends to inquire high-frequent symptoms, while the mutual information matrix can consider a low-frequent symptom, as long as it has a strong connection with a particular disease than the others.  }
\label{fig:mutual}
\end{figure}

\begin{figure*}[t]
\centering
  \includegraphics[width=.95\linewidth]{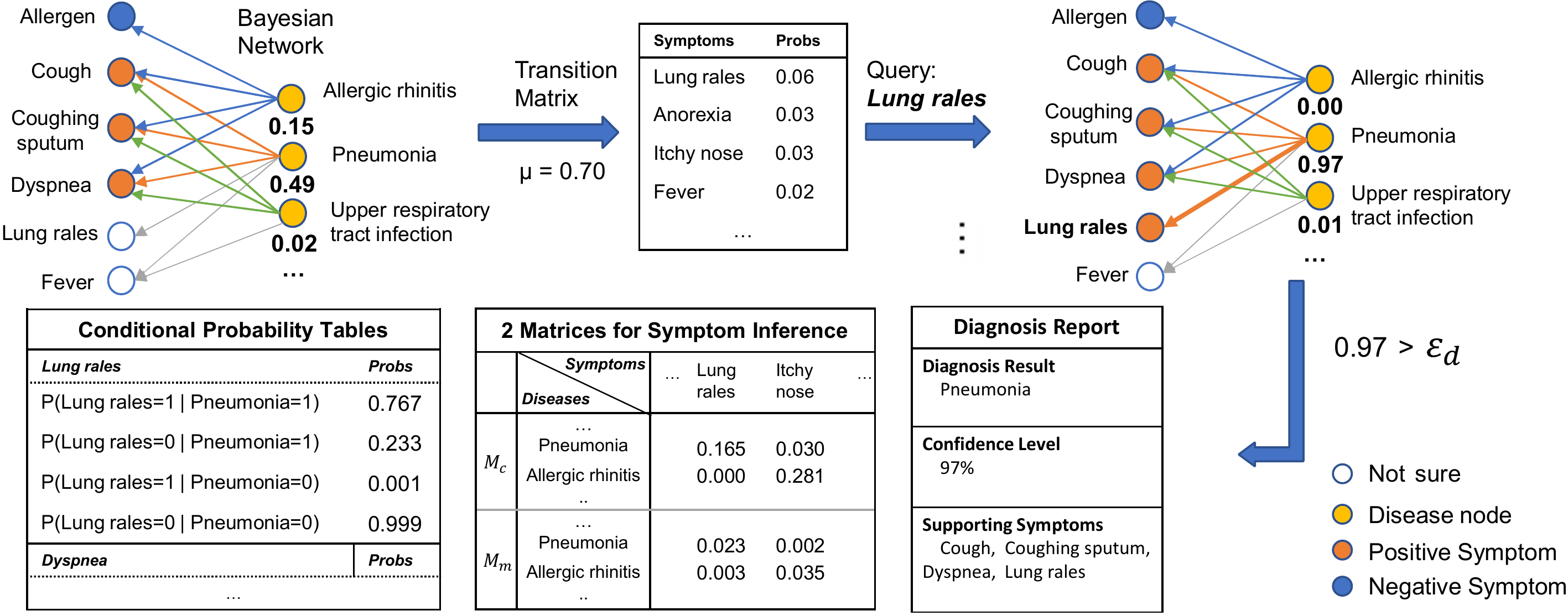}
\caption{Case study of the model parameters and its decision process in  two dialogue turns.}
\vspace{-2mm}
\label{fig:casestudy}
\end{figure*}

\paragraph{Ablation Studies.}
We also conduct ablation studies to test the effectiveness of each component. The results are shown in Table \ref{table:Abl}. 
We first analyze the effectiveness of the two matrices used to model different symptom inquiry logics. Specifically, we conduct the experiments of removing one of them respectively and replacing them both with an MLP.
Then, we further ablate the BayesNet used for disease inference, and the DM module degenerates to a 3-layer MLP that learns with the A2C algorithm. 
The results of ablation study are presented in the last line of Table \ref{table:Abl}. 
We can see that the ablation of every component can all cause a drop in the overall performance, demonstrating the indispensability of each part.

\subsection{Analysis of Interpretability}
During applications, one can easily analyze how BR-Agent arrives at the decision of each turn. 
For instance, we could explain why it queries a particular symptom by investigating its disease estimation at that turn from the BayesNet and its major inquiry logic (to ensure/distinguish) determined by the logic switcher. 
Besides, the parameters of the BayesNet and two transition matrices are all practically meaningful, such as referring to the prior probability of a disease or the conditional probability of a symptom given diseases.  
We could easily look into these values to analyze the model's estimation of them and see whether they are consistent with the clinical guideline.  
Detailed examples and analyses are provided below.

\paragraph{Case Study of the Decision Process.} 
We first present a case study to show how we can utilize the transparency of BR-Agent to analyze how it arrives at a particular action. 
Fig. \ref{fig:casestudy} presents a case study of the BR-Agent's decision process in two dialogue turns. We can see that during symptom inquiry, the weight $\mu$ produced by the logic switcher is equal to 0.7, indicating that the conditional probability matrix dominates the current inquiry logic. 
To ensure its suspicion of the disease \emph{pneumonia}, BR-Agent queries its typical symptom \emph{lung rales}. 
After this turn of inquiry, its estimation of the probability of \emph{pneumonia} increases from 0.49 to 0.97, leading to the final diagnosis. 
As shown by the diagnosis report in the bottom-right corner of Fig. \ref{fig:casestudy}, the confidence level of the diagnosed disease \emph{pneumonia} is 97\%, which is derived from the disease distribution in the final turn; the supporting symptoms for its diagnosis are the ones that are successfully queried (represented as orange nodes in the figure) and connected to the disease node \emph{pneumonia} in the BayesNet at the same time.

\paragraph{Human Evaluation of the Decision Process.} 
Then, we further analyze whether the inference process of BR-Agent is rational from the professional perspective. To this end, we invite two clinical experts to conduct the evaluation. 
For a symptom inquiry action, they score whether it is rational to query the symptom based on its current disease estimation and whether its inquiry logic (to ensure/distinguish) is reasonable.
For a diagnosis action, they evaluate whether it is reasonable to diagnose a disease based on its already-known symptom information.
The score is given on a 5-point Likert scale (5 for the best). 
As the existing DSMD methods are not transparent enough to conduct a similar analysis, we design two baselines by ourselves for comparison, which are Transparent-A2C and a retrieval-based method. 
Transparent-A2C is a transparent variant of A2C-GCN, while the retrieval-based method can explain the model's action by retrieving and analyzing the most similar sample in the dataset. 

The average rationality scores and the Cohen's kappa between annotators are listed in Table~\ref{table:human}. 
We can see that BR-Agent obtains remarkably high rationality scores, 4.30 for the symptom inquiry action and 4.39 for the diagnosis action, while all the scores of the two baselines are no more than 3 points. 
It demonstrates that the actions of BR-Agent are very rational from the professional perspective, as it reasonably mimics real doctors' inquiry logics. The Cohen's kappa are all between 0.69 and 0.82, indicating strong inter-annotator agreement. 

\paragraph{Analysis of the Neural Logic Switcher and Two Matrices.} 
To more clearly illustrate the different symptom inquiry logics of the two matrices in BR-Agent, we present a case study in Fig. \ref{fig:mutual}. 
Before that turn of symptom inquiry, the system has high suspicion of Hand-Foot-and-Mouth Disease (HFMD) and Upper Respiratory Infection (URI). 
We can see that the conditional probability matrix tends to inquire the high-frequent symptom ``fever''. 
However, ``fever'' is a common symptom for both HFMD and URI, so querying ``fever'' would not be able to confirm any one of them.
On the other hand, the mutual information is able to consider the comparatively low-frequent symptom ``herpes'', as it is a typical symptom of HFMD and can be used to effectively distinguish the two diseases.  
In this example, the neural logic switcher generates the $\mu = 0.3$, \footnote{Note that the lower value of $\mu$ indicates stronger dominance of the mutual information matrix in the overall symptom inquiry logic. } choosing the mutual information as the more dominant logic. As a result, the system inquires ``herpes''. 
After confirming this symptom, the system effectively arrives at the diagnosis result of HFMD with the confidence level of 96\%.
By analyzing the distribution of weight $\mu$ generated by the neural logic switcher in the datasets, we find that the average value of $\mu$ is 0.44, and the mutual information matrix is more frequently chosen as the dominant inquiry logic. 

\paragraph{Analysis of Parameter Transparency.}
Some of the parameters in the BayesNet and the two transition matrices are listed in Fig. \ref{fig:mutual} and Fig. \ref{fig:casestudy}.  These parameters are all practically meaningful. For instance, we can see that the occurrence of disease \emph{pneumonia} in Fig. \ref{fig:casestudy} strongly affects the distribution of symptom \emph{lung rales}, which is in accordance with the clinical guideline. However, we also find that many of the Bayesian parameters stays near the value of 0.5, which is the value we use to initialize the parameters that cannot be more estimated based on the dataset (see \Cref{sec:di}). It is probably because the datasets are too small to fine-tune all the parameters.

\section{Conclusion}
In this paper, we made an initial attempt towards the interpretable data-driven DSMD. 
To this end, we proposed a novel method to realize the DM of DSMD with interpretable decision process and transparant components. It consists of a BayesNet for disease inference and two matrices to simulate human doctors' symptom inquiry logics, controlled by a neural logic switcher. 
We also constructed a large DSMD dataset to alleviate the data sparsity issue. 
Empirical results showed that our method exceeded the previous state-of-the-art by a large margin in both diagnosis accuracy and symptom recall. 

\section*{Acknowledgments}
This work was supported by National Natural Science Foundation of China (Grant No.61976233), Guangdong Province Basic and Applied Basic Research (Regional Joint Fund-Key) Grant No.2019B1515120039 and Shenzhen Fundamental Research Program (Project No.RCYX20200714114642083, No.JCYJ20190807154211365). 
It was also supported by the Research Grants Council of Hong Kong (PolyU/15207920, PolyU/15207821) and National Natural Science Foundation of China (62076212).
We also thank Jian Wang, Zijing Ou, and Yueyuan Li
for their helpful comments on this paper.

\clearpage

\bibliographystyle{named}
\bibliography{anthology,custom}

\end{document}